\documentclass[]{ceurart}

\begin{document}

\copyrightyear{2025}
\copyrightclause{Copyright for this paper by its authors.
  Use permitted under Creative Commons License Attribution 4.0
  International (CC BY 4.0).}

\conference{De-Factify 4: 4th Workshop on Multimodal Fact-Checking and Hate Speech Detection, co-located with AAAI 2025. Philadelphia, Pennsylvania, USA}

\title{AI-generated Text Detection: A Multifaceted Approach to Binary and Multiclass Classification}

\author[1]{Harika Abburi}[%
email=abharika@deloitte.com,
]
\address[1]{Deloitte \& Touche Assurance and Enterprise Risk Services India Private Limited, India}
\address[2]{Deloitte \& Touche LLP, USA}

\author[2]{Sanmitra Bhattacharya}[%
email=sanmbhattacharya@deloitte.com,
]

\author[2]{Edward Bowen}[%
email=edbowen@deloitte.com,
]

\author[1]{Nirmala Pudota}[%
email=npudota@deloitte.com,
]

\begin{abstract}
Large Language Models (LLMs) have demonstrated remarkable capabilities in generating text that closely resembles human writing across a wide range of styles and genres. However, such capabilities are prone to potential misuse, such as fake news generation, spam email creation, and misuse in academic assignments. As a result, accurate detection of AI-generated text and identification of the model that generated it are crucial for maintaining the responsible use of LLMs. In this work, we addressed two sub-tasks put forward by the Defactify workshop under AI-Generated Text Detection shared task at the Association for the Advancement of Artificial Intelligence (AAAI 2025): Task A involved distinguishing between human-authored or AI-generated text, while Task B focused on attributing text to its originating language model. For each task, we proposed two neural architectures: an optimized model and a simpler variant. For Task A, the optimized neural architecture achieved fifth place with $F1$ score of 0.994, and for Task B, the simpler neural architecture also ranked fifth place with $F1$ score of 0.627.

\end{abstract}

\maketitle

\section{Introduction}
The domain of Natural Language Generation (NLG) is witnessing a remarkable transformation with the emergence of Large Language Models (LLMs). LLMs, characterized by their large parameter size, have shown state-of-the-art (SOTA) capabilities in generating text that closely mirrors the verbosity and style of human language. They have been shown to outperform traditional Natural Language Processing (NLP) approaches across a wide range of applications, such as story generation \cite{fan2018hierarchical}, Artificial Intelligence-assisted writing \cite{hutson2021robo}, conversational response generation \cite{mousavi2023response}, and code auto-completion \cite{tang2023domain}. 


Although LLMs ability to generate human-like text is impressive, it also presents a growing threat: malicious actors can exploit this capability for unethical purposes, such as fake news generation \cite{uchendu2021turingbench} and social harm \cite{kumar2023language}.
Understanding whether content originates from an AI system or a human is crucial to ensuring appropriate use in downstream applications with suitable oversight. Additionally, knowing the specific LLM responsible for AI-generated text helps in managing potential biases and limitations inherent to that model. Given these concerns, it is essential to advance techniques for the automatic detection of AI-generated content and its attribution to the originating model (model attribution) \cite{wu2025survey,bethany2024deciphering}.

Early efforts in detecting AI-generated text, Gehrmann et al. \cite{gehrmann2019gltr} employed statistical techniques combined with visual tools, capitalizes on the assumption that AI systems tend to rely on limited language patterns with high confidence scores. Other studies \cite{galle2021unsupervised,opara2024styloai,godghase2024distinguishing} used features such as stylometry, lexical analysis, and readability to distinguish between human-authored and AI-generated text.  Following these foundational approaches, zero-shot detection methods have gained a lot of attention, focusing on analyzing LLM outputs through entropy, log-probability scores, and perplexity \cite{su2023detectllm,wu2023llmdet}. For example, Fast-DetectGPT \cite{bao2023fast} proposed a solution by leveraging conditional probability curvature and a sampling strategy. Furthermore, a new method called Binoculars is introduced by \cite{hans2024spotting} for the detection of AI-generated text, which requires only simple calculations using a pair of pre-trained LLMs. The Binoculars score is essentially a normalized version of perplexity, obtained by dividing perplexity by cross-perplexity. Despite their promising results, zero-shot methods are less reliable in scenarios lacking direct access to LLM internals \cite{zhang2023assaying,yang2023survey}. An alternative strategy is to fine-tune pre-trained models on datasets with human-authored and AI-generated texts, improving detection precision and reliability \cite{valiaiev2024detection}. The Ghostbuster \cite{verma2023ghostbuster} proposed an ensemble approach with a series of weaker models, followed by a search over combination functions and then a linear classifier. Abburi et. al., \cite{abburi-etal-2023-simple} proposed an ensemble approach by concatenating probabilities from pre-trained LLMs as feature vectors for machine learning classifiers. They also fine-tuned another model by extracting the document-level representations derived from well established techniques \cite{abburi2024toward, abburi2024team}.

Many techniques utilized for detecting AI-generated content, such as Fast-DetectGPT, Ghostbuster, and various ensemble methods discussed earlier, are also applicable to model attribution. Initially, features including n-gram frequency, lexical analysis, Linguistic Inquiry and Word Count (LIWC), and readability scores are extracted and used as inputs for traditional machine learning and deep learning models \cite{uchendu2020authorship}. With advancements in neural networks, transformer-based models such as Robustly optimized Bidirectional Encoder Representations from Transformers (BERT) approach RoBERTa, BERT, and Decoding- enhanced BERT with disentangled attention (DeBERTa) models are also explored for model attribution \cite{uchendu2021turingbench,li2022artificial,he2023mgtbench}.


To boost this area of research further, the Defactify workshop under AI-Generated Text Detection shared task at AAAI 2025 put forth two sub-tasks: 1. Task A involves binary classification, where the goal is to determine whether a given text is AI-generated or human-authored. 2. Task B focuses on model attribution (multiclass classification), builds on Task A to identify which specific LLM generated a given piece of AI-generated text. For each task, we developed two neural architectures: an optimized model and a simpler architecture. We optimized the neural architecture proposed in \cite{abburi2024toward} by replacing token-level features with stylometry features and extracting document-level representations from three techniques: 
 the RoBERTa-base AI detector, stylometry features, and EmbEddings from bidirEctional Encoder rEpresentations (E5) model. These features are then fed into a fully connected layer to produce final predictions. Additionally, to address the challenges of Task B, we also proposed a simple and efficient gradient boosting classifier with stylometric and SOTA embeddings as features.


\begin{table}[h]
\centering
\caption{Data statistics}
\label{stats}
\begin{tabular}{|p{4cm}|c|c|} 
\hline 
\hspace{0.6cm} \textbf{Category} & \textbf{$Training$} & \textbf{$Validation$}  \\
\hline
Human &7,255&1,569\\
\hline
\hline
Gemma-2-9b & 7,255 & 1,569\\
GPT\_{4}-o & 7,255 & 1,569\\
LLAMA-8B & 7,255 & 1,569\\
Mistral-7B & 7,255 & 1,569\\
Qwen-2-72B & 7,255 & 1,569\\
Yi-large & 7,255 & 1,569\\
\hline
Total AI samples & 43,530& 9,414 \\
\hline
\hline
\hspace{0.5cm} Human + AI samples & 50,785& 10,983 \\
\hline
\end{tabular}
\end{table}

\section{Dataset}
The dataset used in this study is provided by the shared task organizers\footnote{https://defactify.com/ai\_gen\_txt\_detection.html}. Table \ref{stats} shows the number of human-authored (Human) and AI-generated (AI) samples available in both the training and validation splits for each category in the dataset. The dataset is structured such that each prompt includes a human-authored story accompanied by parallel AI-generated text from various LLMs. For Task A (binary classification), the categories include Human and AI, while Task B (multiclass classification) expands to seven categories: Gemma-2-7B, GPT\_4-o, Human, LLAMA-8B, Mistral-7B, Qwen-2-72B, and Yi-large, as detailed in Table \ref{stats}.

The shared task organizers evaluate the performance of the submissions using a test set comprising 10,963 samples. The exact distribution of samples for each category is not disclosed to participants during the results submission phase. Additional details about the task and dataset are available at \cite{roy-2025-defactify-overview-text, roy-2025-defactify-dataset-text}.

\section{Approach}
Our approach extends the neural architecture proposed by \cite{abburi2024toward}, which was initially designed for binary classification to distinguish between human-authored and AI-generated text. We expanded this architecture to address both binary and multiclass classification tasks, with the latter including seven categories, such as different LLMs and human-authored text.


The original fine-tuning based architecture \cite{abburi2024toward} extracts document-level representations from three main components: First, it employs a pre-trained RoBERTa-base AI detector \cite{solaiman2019release} which is foundation model for distinguishing between AI-generated and human-authored text. Second, a Bidirectional Long Short-Term Memory (BiLSTM) attention layer processes token-level perplexity (log-probability of the observed token, log-probability of the most likely token according to the model, and entropy of the token probability distribution at a given position) and word based frequency features which are extracted from variants of GPT2 model. Third, the SOTA EmbEddings from bidirEctional Encoder rEpresentations (E5) model \cite{wang2022text} is utilized to enhance semantic understanding across texts. Each of these representations are concatenated and fed into a fully connected layer to produce final predictions. Hereafter, we refer to this comprehensive architecture as the \emph{Full Architecture}.

\begin{figure}[h]
\centering
\includegraphics[width=0.9\textwidth]{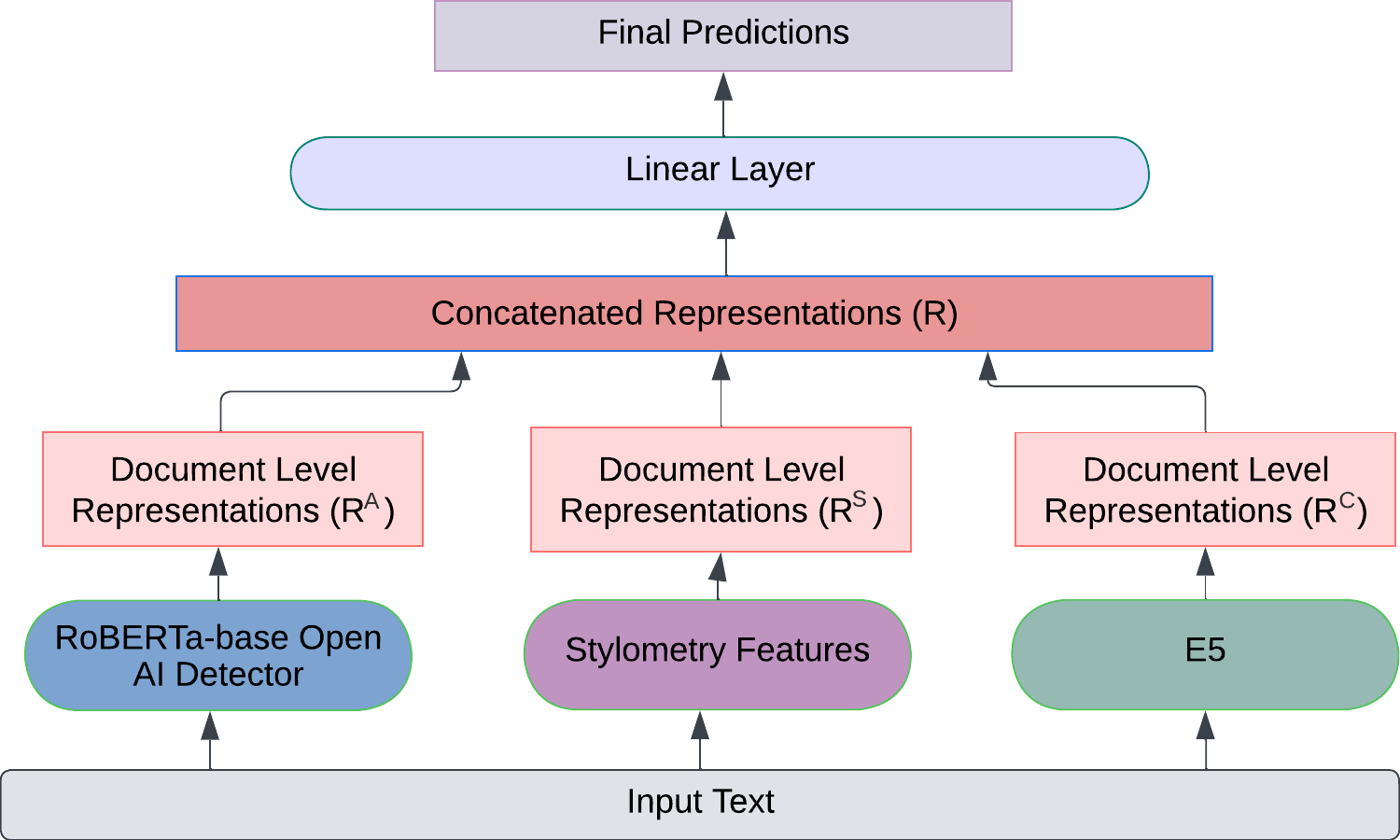}%
\caption{Proposed optimized neural architecture}\label{fig:PA}
\end{figure}

In addition to experimenting with \emph{Full Architecture} proposed in \cite{abburi2024toward}, we also explored various combinations of the same architecture to simplify its complexity. Notably, an architecture incorporating only the RoBERTa base OpenAI detector and E5 achieved performance comparable to \emph{Full Architecture}. Additionally, substituting token-level features with stylometry features (as discussed in subsection \ref{sty}) demonstrated enhanced performance for both sub-tasks. This modification allowed us to optimize the \emph{Full Architecture} by reducing its complexity while maintaining high performance. We henceforth refer to this optimized version as \emph{Optimized Architecture}, as shown in Figure \ref{fig:PA}. For both architectures, the hyperparameter settings align with those of the original framework by \citet{abburi2024toward}.

\subsection{Stylometry Features}
\label{sty}
Incorporating stylometric features \cite{opara2024styloai, godghase2024distinguishing} into our \emph{Optimized Architecture} plays a crucial role in improving text predictability and distinguishing between human-authored and AI-generated text, as well as text produced by different LLMs. Stylometry examines distinct linguistic and structural characteristics within a text, providing valuable insights into an author's unique style or the underlying properties of the text.


Building on the feature importance highlighted by \cite{opara2024styloai}, we selected a set of eight features: unique word count, stop word count, moving average type-token ratio (MTTR), hapax legomenon rate, word count, bigram uniqueness, sentence count, and average sentence length. Furthermore, we also added three more features identified by \cite{godghase2024distinguishing} such as lowercase letter ratio, burstiness, and verb ratio for their ability to capture distinct stylistic elements and text dynamics, resulting in a total of 11 features. These features collectively provide an understanding of the stylistic nuances inherent in different text sources. By leveraging these features, our architecture is better equipped to detect subtle variations and discriminate between multiple text categories, enhancing its overall effectiveness.


\subsection{Simple Architecture}
In our attempts to apply the \emph{Optimized Architecture} to a multiclass setup, we encountered a few challenges. The RoBERTa-base AI detector, originally fine-tuned for binary classification, did not effectively extend to the multiclass classification task. This limitation prompted us to explore alternative approaches that are better suited to distinguish among the seven categories, including various LLM-generated texts and human-authored content.

\begin{figure}[h]
\centering
\includegraphics[width=0.5\textwidth]{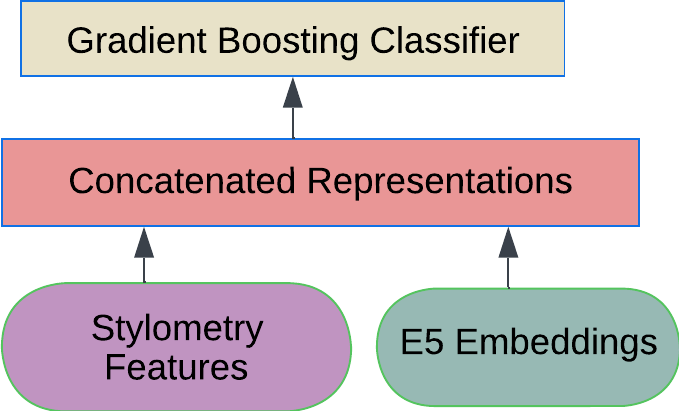}%
\caption{Simple architecture}\label{fig:gb}
\end{figure}

To address these challenges, we proposed a \emph{Simple Architecture} (as shown in Figure \ref{fig:gb}) that combines the strengths of semantic representation and stylistic analysis. We concatenated the robust semantic embeddings from the E5 model with 11 carefully selected stylometric features. This combination captures both the deep semantic content and the subtle stylistic patterns present in the text. By inputting this enriched representation into a gradient boosting classifier, we aimed to enhance the model's ability to perform fine-tuned classification across multiple classes. By taking these steps, we enhance our architecture’s capability to correctly classify and distinguish between diverse text sources, leading to more reliable results.


\section{Results}
We evaluated three distinct approaches: the \emph{Full Architecture}, the \emph{Optimized Architecture}, and the \emph{Simple Architecture}, across binary and multiclass classification tasks. Each architecture exhibited individual strengths and limitations, providing valuable insights into the effectiveness of different architectures. On the validation set, all architectures achieved an $F1$ score of 1.0 for Task A and 0.9 for Task B. Therefore, we are not showing results on the validation set and only discuss the results on the test set.

Table \ref{result} presents the $F1$ scores with different architectures across both tasks on the test set. The \emph{Full Architecture} served as our benchmark, achieving a strong $F1$ score of 0.949 in Task A. However, its performance significantly dropped in the complex multiclass classification setting of Task B, with a score of only 0.19.


The \emph{Optimized Architecture}, which incorporated stylometric features and reduced complexity, outperformed the \emph{Full Architecture} with $F1$ scores of 0.994 in Task A and 0.406 in Task B. Notably, this architecture excelled in Task A binary classification, showcasing its strength in detecting AI-generated text. However, the multiclass classification in Task B remains a challenge, highlighting areas that require further improvement.


\begin{table}[!h]
\centering
\caption{$F1$ scores for different architecture's across both tasks}
\label{result}
\begin{tabular}{|p{4cm}|c|c|} 
\hline 
\hspace{0.6cm} \textbf{Architecture's} & \textbf{$Task A$} & \textbf{$Task B$}  \\
\hline
\emph{Full Architecture} &0.949&0.190\\
\hline
\emph{Optimized Architecture} &\textbf{0.994}&0.406\\
\hline
\emph{Simple Architecture} &0.974&\textbf{0.627}\\
\hline
\end{tabular}
\end{table}

To address these challenges, we implemented the \emph{Simple Architecture}, leveraging the E5 semantic embeddings and stylometry features with gradient boosting classifier. The outcome of this architecture led to a notable 22\% improvement in Task B performance compared to the \emph{Optimized Architecture}. However, it resulted in a slight 2\% decrease in Task A performance.

\section{Conclusion}
In this paper, we described our submission to the AI-Generated Text Detection shared task at AAAI 2025 which includes binary and multiclass classification of generative AI content. To tackle both these tasks, we proposed an optimized and a simple architecture. Our experiments demonstrated that the optimized architecture is a promising approach; it secured fifth place with a $F1$ score of 0.994 for the binary classification task. Similarly, utilizing simple architecture for the multiclass classification task, we also achieved fifth place with a $F1$ score of 0.627. While our approach shows promising results for the binary task, further exploration is needed to enhance and refine our models for the multiclass classification task.

\bibliography{sample-ceur}
\end{document}